\documentclass[lettersize,journal]{IEEEtran}
\usepackage{amsmath,amsfonts}
\usepackage{algorithmic}
\usepackage{array}
\usepackage{textcomp}
\usepackage{stfloats}
\usepackage{url}
\usepackage{verbatim}
\usepackage{graphicx}

\usepackage[table]{xcolor}
\definecolor{light-gray}{gray}{0.9}
\usepackage{amsmath}
\usepackage{amssymb}
\usepackage{booktabs}
\usepackage{multirow}

\usepackage{algorithm}
\usepackage{cite}
\usepackage{multirow}
\usepackage{tabularx}
\usepackage{booktabs}
\usepackage{bbm}
\usepackage{amssymb}
\usepackage{subcaption}
\usepackage{stfloats}
\usepackage[pagebackref,breaklinks,colorlinks]{hyperref}

\usepackage{newfloat}
\usepackage{listings}
\floatstyle{ruled}
\newfloat{listing}{tb}{lst}{}
\floatname{listing}{Listing}

\def\BibTeX{{\rm B\kern-.05em{\sc i\kern-.025em b}\kern-.08em
    T\kern-.1667em\lower.7ex\hbox{E}\kern-.125emX}}
\usepackage{balance}
\begin{document}
\title{\huge Adapt Anything: Tailor Any Image Classifiers across Domains And Categories Using Text-to-Image Diffusion Models}
\author{Weijie Chen, Haoyu Wang, Shicai Yang, Lei Zhang, Wei Wei, Yanning Zhang\\
Luojun Lin, Di Xie, Yueting Zhuang,~\IEEEmembership{Senior Member,~IEEE}
\thanks{W. Chen and Y. Zhuang are with the College of Computer Science and Technology, Zhejiang University, Hangzhou, China. Email: {\small \{chenweijie, yzhuang\}@zju.edu.cn}}\protect
\thanks{S. Yang and D. Xie are with Hikvision Research Institute, Hangzhou, China. Email: {\small \{yangshicai, xiedi\}@hikvision.com}}\protect
\thanks{H. Wang is with Northwestern Polytechnical University, Xi'an, China. Work was done when he was an intern in Hikvision Research Institute. Email: {\small wanghaoyunwpu@mail.nwpu.edu.cn}}\protect
\thanks{L. Zhang, W. Wei, and Y. Zhang are with Northwestern Polytechnical University, Xi'an, China. Email: {\small \{nwpuzhanglei, weiweinwpu, ynzhang\}@nwpu.edu.cn}}\protect
\thanks{L. Lin is with the College of Computer and Data Science, Fuzhou University, Fuzhou, China. Email: \small linluojun2009@126.com}\protect
\thanks{L. Lin, D. Xie, and Y. Zhuang are the corresponding authors.}
}

\markboth{Journal of \LaTeX\ Class Files,~Vol.~18, No.~9, September~2020}%
{How to Use the IEEEtran \LaTeX \ Templates}

\maketitle

\begin{abstract}
We do not pursue a novel method in this paper, but aim to study if a modern text-to-image diffusion model can tailor any task-adaptive image classifier across domains and categories. Existing domain adaptive image classification works exploit both source and target data for domain alignment so as to transfer the knowledge learned from the labeled source data to the unlabeled target data. However, as the development of the text-to-image diffusion model, we wonder if the high-fidelity synthetic data from the text-to-image generator can serve as a surrogate of the source data in real world. In this way, we do not need to collect and annotate the source data for each domain adaptation task in a one-for-one manner. Instead, we utilize only one off-the-shelf text-to-image model to synthesize images with category labels derived from the corresponding text prompts, and then leverage the surrogate data as a bridge to transfer the knowledge embedded in the task-agnostic text-to-image generator to the task-oriented image classifier via domain adaptation. Such a one-for-all adaptation paradigm allows us to adapt anything in the world using only one text-to-image generator as well as the corresponding unlabeled target data. Extensive experiments validate the feasibility of the proposed idea, which even surpasses the state-of-the-art domain adaptation works using the source data collected and annotated in real world.
\end{abstract}


\section{Introduction}
\IEEEPARstart{G}{eneralization} and adaptation are two very fundamental research topics in machine learning. In comparison, generalization urges the model to generalize to the unseen target data \cite{Zhou2021DomainGW,Meng2022AttentionDF,Sun2022DynamicDG}, while adaptation allows the model to get evolved using the seen target data without annotations \cite{Ganin2014UnsupervisedDA,Chen2022LearningDA,Lang2023ExploringID}. Adaptation is a very crucial supplementary ability for generalization. It is inevitable that the model cannot perform well in every target domain due to the limited generalization ability. As a result, the model is expected to be adaptive to each target data via the unsupervised optimization in the target domain, which drives a population of researchers delve into unsupervised domain adaptation (UDA) in the past few years.

\begin{table}[t]
\centering
\caption{Prior UDA methods are designed in a one-for-one learning paradigm, which exploit task-specific source knowledge for target adaptation, such as conventional UDA \cite{Ganin2014UnsupervisedDA}, source-free UDA \cite{Chen2021SelfSupervisedNL}, and syn-to-real adaptation methods \cite{Ganin2014UnsupervisedDA}. For the purpose of AGI, we propose Adapt Anything as a one-for-all adaptation paradigm in this paper. Only one text-to-image generator is all we need to transfer the knowledge to different target tasks across domains and categories.}
\label{table::task-comparison}
        \setlength\tabcolsep{4pt}
	\resizebox{.48\textwidth}{!}{
	\begin{tabular}{l|l|c}
	\hline
	Problem Setting& Source Knowledge & Adaptation Paradigm\\
	\midrule
	Conventional UDA & Labeled Source Data (Task-Specific) & One-for-One \\
	\midrule
	Source-Free UDA & Pre-trained Source Model (Task-Specific) & One-for-One \\
        \midrule
        Syn-to-Real Adaptation & 3D Model for Rendering (Task-Specific) & One-for-One \\
	\midrule
        \rowcolor{light-gray}
	Adapt Anything (Ours) & Text-to-Image Generator (\textbf{Task-Agnostic}) & \textbf{One-for-All} \\
	\hline
	\end{tabular}
	}
\end{table}

\begin{figure}[t]
  \centering
  \includegraphics[width=1.\linewidth]{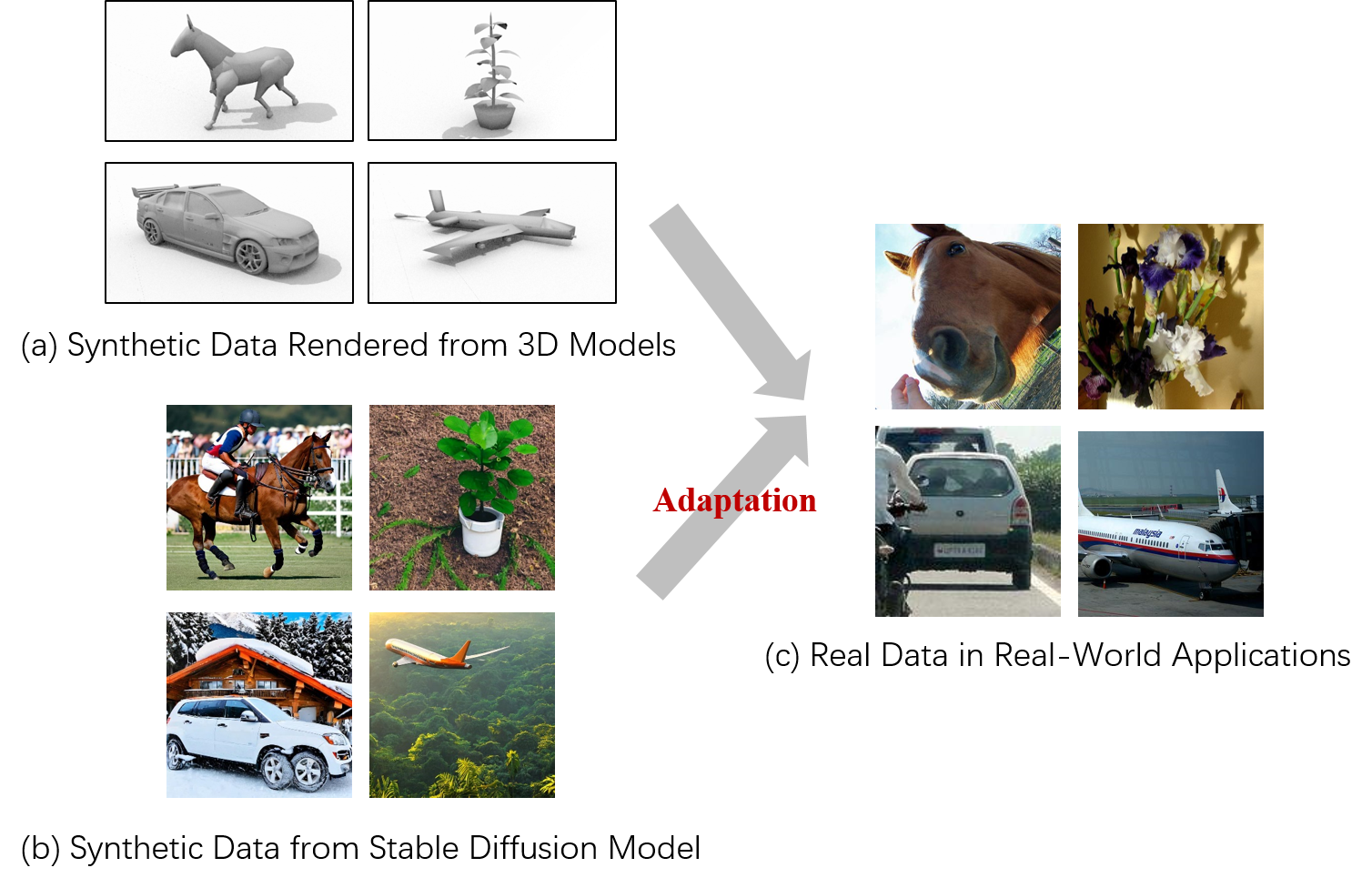}
  \caption{Syn-to-real adaptation aims to synthesize fake images as source data and propagate the knowledge to the unlabeled target domain in the real-world application. Previous works usually synthesize data using 3D models from different angles and with different lighting conditions, such as VisDA17~\cite{Peng2017VisDATV}. In this paper, we use a general text-to-image diffusion model to synthesize surrogate source data without designing different 3D models for task-specific visual categories.}
  \label{fig:task}
\end{figure}

UDA is defined as a process of knowledge transfer from source knowledge to the unlabeled target data. By taking different variants of source knowledge into consideration, UDA is developed into different variants. As listed in Table \ref{table::task-comparison}, conventional UDA annotates the source data collected in real world, and then apply domain alignment to transfer the knowledge from the labeled source data to the unlabeled target data \cite{Ganin2014UnsupervisedDA,Saito2017MaximumCD}. In some specific applications, for the purpose of data privacy protection, source-free UDA is came up which exploits a source pre-trained model as source knowledge instead of source data, and then utilizes the source model for pseudo labeling and self-training in target data \cite{Chen2021SelfSupervisedNL,Liang2020DoWR}. These two UDA variants use source data directly or indirectly. Unfortunately, source data is expensive and difficult to collect and annotate in real world. To tackle this challenge, syn-to-real adaptation is proposed, which, for example, uses 3D models or game engines as source knowledge, synthesizing virtual images as a surrogate of source data to conduct knowledge transfer \cite{Peng2017VisDATV,Peng2018Syn2RealAN}, as shown in Figure \ref{fig:task}. However, the aforementioned UDA methods are designed as a one-for-one adaptation paradigm, since the source knowledge is task-specific. We have to construct different source knowledge for different target tasks, which violates the final purpose of Artificial General Intelligence (AGI). For instance, if we want to build a vehicle recognition model adapted to different weather conditions or other agnostic target domains, a preliminary source knowledge about vehicle is necessary at least, so are other countless target tasks. Fortunately, as the development of the remarkable text-to-image diffusion model, which can synthesize high-fidelity images given task description using text prompts, we wonder if the synthetic data from text-to-image generator can serve as a surrogate of source knowledge. If this is OK, only one remarkable text-to-image generator is all we need to transfer the knowledge to every target task across different domains and categories in a one-for-all manner, aka Adapt Anything we propose in this paper. We try to answer this question by taking image classification task as an example.

\begin{figure}[t]
    \centering
    \includegraphics[width=0.45\linewidth]{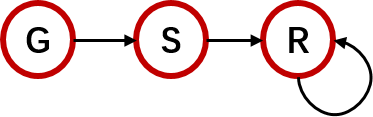}
    \caption{Knowledge flow. ``G'' is a pre-trained text-to-image generator. ``S'' is the synthetic data with labels derived from text prompts. ``R'' is the unlabeled real data in a target domain. The arrows denote three stages of knowledge flow: 1) The knowledge in the text-to-image generator is saved as the synthetic data; 2) The synthetic data is served as a bridge to propagate the initial knowledge from the generator to the classifier in target data; 3) Given the initial knowledge, self-training is further utilized to conduct model self-evolution in target data to achieve better adaptation performance.}
    \label{fig:knowledge-flow}
\end{figure}

\begin{figure*}[htbp]
  \centering
  \includegraphics[width=0.945\linewidth]{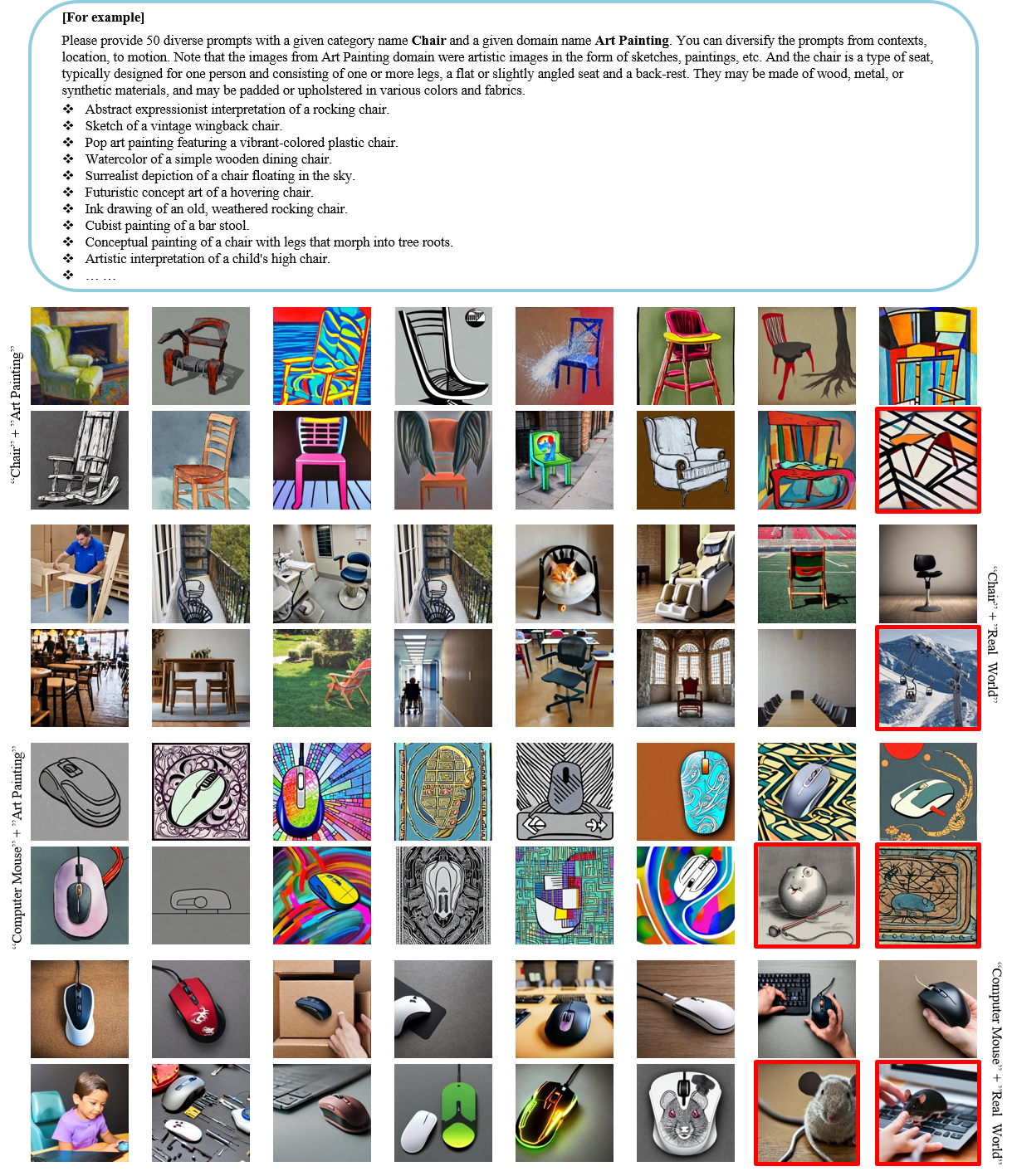}
  \caption{The process of image synthesis. The upper box is an example of the dialogue with ChatGPT. We provide the detailed description about category and domain information, and ask ChatGPT to provide diverse text prompts so as to drive the Stable Diffusion to generate diverse synthetic data. The bottom figures present the synthesized results about the categories ``Chair'' and ``Computer Mouse'' in the domains of ``Art Painting'' and ``Real World''. Although most images are matched to what we want, there still exist some semantic-mismatched images, such as the ones with {\color{red}{\textbf{red}}} border. For example, the model sometimes cannot understand ``Computer Mouse'' and split this phrase into ``Computer'' and ``Animal Mouse'' to drive image synthesis, which can be viewed as noises in the synthetic data. This is exactly the motivation we propose to discard the synthetic domain and conduct intra-domain knowledge transfer in the third stage. (Best viewed in color.)}
  \label{fig:data}
\end{figure*}

To build the framework of Adapt Anything, three vanilla stages are necessary. The overall knowledge flow is illustrated in Figure \ref{fig:knowledge-flow}: 1) \textbf{Synthetic data generation}. With the given task description, including domain and category information in target data, we can synthesize abundant surrogate source data by feeding text prompts to a text-to-image generator. Without specific statement, we use Stable Diffusion \cite{rombach2021highresolution} by default in this paper, which is a state-of-the-art text-to-image generator. However, it will lead to mode collapse of image generation if using similar text prompts. To increase the diversity of image generation, we use ChatGPT to diversify text prompts by providing the detailed domain and category information. In this way, we can synthesize diverse surrogate data so as to dig out the source knowledge embedded in the pre-trained text-to-image generator thoroughly.  2) \textbf{Inter-Domain Knowledge Transfer}. Identical to the existing syn-to-real adaptation methods, many popular state-of-the-art UDA methods can be applied in this stage so as to transfer the knowledge from the surrogate source data to the unlabeled target data. Note that the labels of the synthetic data are derived from the corresponding text prompts without any human annotation. 3) \textbf{Intra-Domain Knowledge Transfer}. As shown in Figure \ref{fig:data}, some of the synthetic images used in the second stage are inevitable to mismatch with the target categories due to the inherent defects of Stable Diffusion or the misunderstanding text prompts. To reduce the effect of negative transfer, we propose to discard the synthetic data in the third stage, and use target data only for self-training. Since the second stage can provide pseudo labels for the target data, we divide the target data into a confident sub-domain and a unconfident sub-domain according to the prediction confidence. We then treat the confident sub-domain as a labeled part and the unconfident sub-domain as an unlabeled part so as to apply the popular semi-supervised learning methods for model self-evolution in the target domain.

Extensive experiments and ablation studies are conducted on four popular domain adaptive image classification datasets, including Office-31 \cite{Saenko2010AdaptingOffice31}, ImageCLEF-DA \cite{Caputo2014ImageCLEF}, Office-Home \cite{Venkateswara2017DeepOfficeHome}, VisDA-17 \cite{Peng2017VisDATV}. Supported by these experimental results, our finding is that \emph{a modern text-to-image diffusion model can indeed tailor any task-adaptive image classifier across domains and categories}. Specifically, we can even achieve superior results on ImageCLEF-DA, Office-Home, and VisDA to those state-of-the-art UDA methods using source data collected in real world or using synthetic data rendered by 3D models. This finding reflects again that AI-Generated Content (AIGC) is indeed an important direction to develop AGI.

\section{Related Works}
\paragraph{Text-to-Image Generation}
Text-to-image generation has made remarkable progress recently with the development of various generative models including GANs (Generative Adversarial Networks) \cite{2014Generative}, auto-regressive models \cite{2017Attention}, and diffusion models \cite{nokey}. GANs were first proposed by Reed \emph{et al.} \cite{2016Generative}, but they suffer from some limitations such as mode collapse and training instability. To address these issues, recent research has looked to Transformer-based auto-regressive models \cite{2021CogView} combined with a method called discrete VAE (Variational Autoencoder) for image tokenization and Transformers for modeling the joint distribution of textual and image tokens \cite{2021Taming}. Diffusion models have also been widely applied to text-to-image generation recently, with works such as GLIDE \cite{Nichol2021GLIDETP}, making innovative use of these models for text-to-image generation by conditioning the model on an input caption. Nichol \emph{et al.} proposed GLIDE \cite{Nichol2021GLIDETP}, which extends earlier diffusion models by allowing textual input to condition the model, and Ramesh \emph{et al.} took this a step further with DALL-E 2 \cite{2022Hierarchical}, which augments the GLIDE model with a supplemental CLIP \cite{2021Learning} image embedding to increase diversity. Other works such as Stable Diffusion \cite{nokey} have emphasized computational efficiency by representing input images as low-dimension latent codes. In this paper, we do not aim to improve text-to-image generators, but to study the applications of the modern text-to-image generators in the syn-to-real domain adaptation tasks.

\paragraph{Unsupervised Domain Adaptation}
Unsupervised Domain Adaptation (UDA) refers to the task of transferring knowledge from a labeled source domain to an unlabeled target domain by training a model with both source and target data. The main focus is to align the distributions between the source and target domains using discrepancy-based methods, typically achieved through the development of different distance metrics \cite{MMD}. However, it was later discovered that pre-defined distance functions were not sufficient to capture the true distance between the source and target domains. Therefore, adversarial training methods were proposed to automatically align the features between the source and target domains. These methods can be classified into three categories: domain-invariant feature learning \cite{Ganin2014UnsupervisedDA}, task-oriented feature learning \cite{Saito2017MaximumCD}, and feature transferability and discriminability balance learning \cite{huang2022balancing}. Additionally, generative adversarial networks (GANs) have also been applied in UDA for domain mapping, such as mapping source images to target styles \cite{CoGAN} or vice versa \cite{PixelDA,CYCADA}. Recently, there has been increasing interest in applying vision transformers to solve UDA problems. For example, Xu \emph{et al.} \cite{Xu2021CDTransCT} propose a weight-sharing triple-branch transformer that utilizes the self-attention and cross-attention mechanisms for both feature learning and domain alignment. Despite the promising results of the existing UDA methods, they heavily depend on the pre-collected and the pre-annotated source data in a one-for-one manner. In this paper, we discuss if the synthetic data from an off-the-shelf text-to-image generator can take place of the source data in real world. If the answer is yes, all we need to adapt anything in the world is a powerful text-to-image generator, which is a one-for-all manner and helps us get close to the goal of AGI.

\paragraph{Semi-Supervised Learning}
The semi-supervised learning (SemiSL) paradigm is designed to leverage knowledge from unlabeled data with the goal of improving model performance. One effective SemiSL approach is pseudo labeling, which selects the most confidently predicted class as pseudo labels. This approach typically relies on entropy minimization to ensure that the pseudo labels are reliable enough~\cite{grandvalet2005semi}. Another popular SemiSL method, known as consistency regularization, is based on the assumption that the model response should be consistent after perturbations are applied to the input. These perturbations can be data augmentation techniques~\cite{Berthelot2019MixMatchAH} or adversarial transformations of input data~\cite{VAT}. Consistency regularization can also be extended to perturb the model itself, such as time ensembling of the model at different time steps with an exponential moving average of the model parameters~\cite{Temporal}, or adversarial perturbations of the model parameters~\cite{VAT}. Also, some works combine pseudo labeling and consistency regularization for SemiSL, including MixMatch~\cite{Berthelot2019MixMatchAH} and FixMatch~\cite{Sohn2020FixMatchSS}. In this paper, we split the target domain into a labeled part and an unlabeled part, and then use SemiSL methods to implement Intra-Domain Knowledge Transfer.

\section{Adapt Anything}

\begin{figure*}[t]
  \centering
  \includegraphics[width=1.\linewidth]{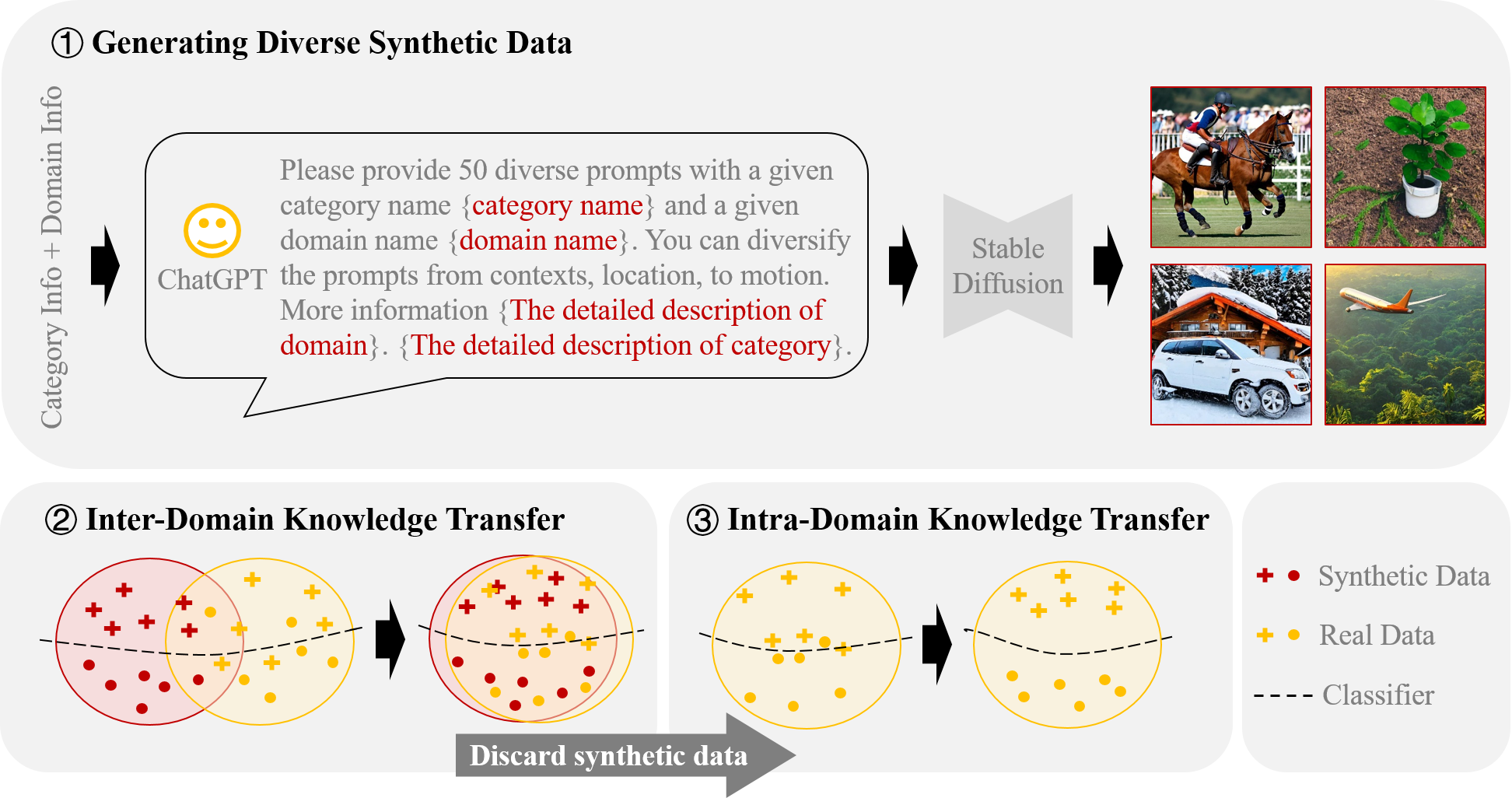}
  \caption{Pipeline of Adapt Anything. 1) Generating diverse synthetic data. Given task information, including category and domain information, we can synthesize task-aware images using text-to-image diffusion model for syn-to-real adaptation. In order to generate diverse synthetic images, we utilize ChatGPT for prompt diversification. 2) Inter-domain knowledge transfer. We can use the popular UDA methods to transfer the knowledge from the synthetic domain to real domain. 3) Intra-domain knowledge transfer. Considering there inevitably exist wrong images during synthetic data generation, we propose to discard the synthetic data. We split the target domain into a confident sub-domain and an unconfident sub-domain, and then apply semi-supervised learning for intra-domain model optimization.}
  \label{fig:pipeline}
\end{figure*}
\subsection{Pipeline}
As shown in Figure \ref{fig:pipeline}, the preliminary stage is to generate diverse synthetic data so as to dig out more source knowledge from a task-agnostic text-to-image Stable Diffusion model. Considering the existence of the inevitable erroneous synthetic data, we propose a simple yet effective \textbf{Coarse-to-Fine} knowledge transfer framework, which is composed of an Inter-Domain Knowledge Transfer stage using the synthetic data and an Intra-Domain Knowledge Transfer stage discarding the synthetic data. The former stage trains an image classifier coarsely by transferring the source knowledge to the unlabeled target data. The latter stage discards the synthetic data and further optimizes the model finely using the target data only for self-training.

\renewcommand\arraystretch{1} 
\begin{table*}[t]
	\centering
        \setlength\tabcolsep{16pt}
	\resizebox{1.\textwidth}{!}{
	\begin{tabular}{l|c|cc|cc|cc|c}
	\hline
        Methods & Real & A$\rightarrow$D & W$\rightarrow$D & D$\rightarrow$A & W$\rightarrow$A & A$\rightarrow$W & D$\rightarrow$W & AVG \\
        \midrule
        DANN \cite{Ganin2014UnsupervisedDA} & \multirow{7}{*}{\checkmark} &79.7 &99.1 &68.2 &67.4 &82.0 &96.9 &82.2 \\
        JAN-A \cite{Long2016DeepTL} & &85.1 &99.7 &69.2 &70.7 &86.0 &96.7 &84.6 \\
        MADA \cite{Pei2018MultiAdversarialDA} & &87.8 & 99.6& 70.3 & 66.4 & 90.0 & 97.4 & 85.2\\
        iCAN \cite{Zhang2018CollaborativeAA}& & 90.1 &\textbf{100.} &72.1 &69.9 &92.5 &98.8 &87.2\\
        CDAN+E \cite{Long2017ConditionalAD} & & 92.9 & \textbf{100.} & 71.0 & 69.3 & 94.1 & 98.6 & 87.7\\
        SymNets \cite{Zhang2019DomainSymmetricNF} & &93.9 &\textbf{100.} &74.6 &72.5 &90.8 &98.8 &88.4 \\
        CDTrans-S \cite{Xu2021CDTransCT} & & 94.6 &99.6 &78.4 &78.0 &93.5 &98.2 &90.4 \\
        CDTrans-B \cite{Xu2021CDTransCT} & & 97.0 &\textbf{100.} &81.1 &81.9 &96.7 &\textbf{99.4} & \textbf{92.6} \\
        \hline
        \hline
        Methods & Real & \multicolumn{2}{c|}{A+W$\rightarrow$D} & \multicolumn{2}{c|}{D+W$\rightarrow$A} & \multicolumn{2}{c|}{A+D$\rightarrow$W} & AVG\\
        \midrule
        DAN \cite{Long2019TransferableRL} & \multirow{4}{*}{\checkmark}  &\multicolumn{2}{c|}{99.6} &\multicolumn{2}{c|}{67.6} & \multicolumn{2}{c|}{97.8} &88.3 \\
        DANN \cite{Ganin2014UnsupervisedDA} & &\multicolumn{2}{c|}{99.7} &\multicolumn{2}{c|}{67.7} & \multicolumn{2}{c|}{98.1} & 88.5\\
        DSBN \cite{Chang2019DomainSpecificBN} & &\multicolumn{2}{c|}{99.0} &\multicolumn{2}{c|}{70.1} & \multicolumn{2}{c|}{98.8} & 89.3\\
        DSAN \cite{Zhu2020DeepSA} & &\multicolumn{2}{c|}{99.1} &\multicolumn{2}{c|}{72.4} & \multicolumn{2}{c|}{98.6} &90.0 \\
        \hline
        \hline
        Methods & Real & \multicolumn{2}{c|}{Syn$\rightarrow$D} & \multicolumn{2}{c|}{Syn$\rightarrow$A} & \multicolumn{2}{c|}{Syn$\rightarrow$W} & AVG\\
        \midrule
         Adapt Anything (Ours)&\multirow{1}{*}{$\times$} &\multicolumn{2}{c|}{95.2} &\multicolumn{2}{c|}{\textbf{82.1}} & \multicolumn{2}{c|}{94.2} & 90.5\\
	\hline
	\end{tabular}
        }
	\caption{Performance comparison with single-source and multi-source UDA using real source data on the Office-31 dataset. ``Real'' means that the labeled source data are collected and annotated in real world. ``$\rightarrow$'' is the knowledge transfer direction. The letters on the left and the right of ``$\rightarrow$'' denotes the source and target domains, respectively. For example, A is the source domain and D is the target domain in ``A$\rightarrow$D''. The same in the following tables.}
	\label{table::results-office31}
\end{table*}

\renewcommand\arraystretch{1} 
\begin{table*}[t]
	\centering
        \setlength\tabcolsep{18pt}
	\resizebox{1.\textwidth}{!}{
	\begin{tabular}{l|c|cc|cc|cc|c}
	\hline
        Methods & Real & I$\rightarrow$P & C$\rightarrow$P & P$\rightarrow$I & C$\rightarrow$I & I$\rightarrow$C & P$\rightarrow$C & AVG \\
        \midrule
        MADA \cite{Pei2018MultiAdversarialDA} & \multirow{12}{*}{\checkmark} & 75.0 & 75.2 & 87.9 & 88.8 & 96.0 & 92.2 & 85.8\\
        iCAN \cite{Zhang2018CollaborativeAA} & & 79.5 & 78.5 & 89.7 & 89.9 & 94.7 & 92.0 & 87.4\\
        CDAN+E \cite{Long2017ConditionalAD} & & 77.7 & 74.2 & 90.7 & 91.3 & 97.7 & 94.3 & 87.7 \\
        CADA-P \cite{Kurmi2019AttendingTD} & & 78.0 & 77.2 & 90.5 & 92.0 & 96.7 & 95.5 & 88.3\\
        MEDM \cite{Wu2020EntropyMV} & & 78.5 & 77.2 & 93.0 & 92.8 & 96.1 & 95.5 & 88.9\\
        MDDA \cite{Wang2019TransferLW} & & 79.8 & 78.8 & 91.5 & 92.0 & 95.7 & 95.5 & 88.9\\
        DADA \cite{Du2020DualAD} & & 79.0 & 77.8 & 93.2 & 92.3 & 98.2 & 95.0 & 89.3\\
        SAFN+ENT \cite{Xu2018LargerNM} & & 80.2 & 78.4 & 93.8 & 92.8 & 96.7 & 95.7 & 89.6\\
        SymNets \cite{Zhang2019DomainSymmetricNF} & & 80.2 & 78.7 & 93.6 & 93.4 & 97.0 & 96.4 & 89.9\\
        DFA-SAFN \cite{Wang2020DiscriminativeFA} & & 80.0 & 78.7 & 94.2 & 93.8 & 97.5 & 96.7 & 90.2\\
        SPL \cite{Wang2019UnsupervisedDA} & & 78.3 & 80.5 & 94.5 & 95.7 & 96.7 & 96.3 & 90.3\\
        MCC+NWD \cite{Chen2022ReusingTT} & & 79.8 & 80.0 & 94.5 & 94.2 & 98.0 & 97.5 & 90.7\\
	\hline
	\hline
        Methods & Real & \multicolumn{2}{c|}{Syn$\rightarrow$P} & \multicolumn{2}{c|}{Syn$\rightarrow$I} & \multicolumn{2}{c|}{Syn$\rightarrow$C} & AVG\\
        \midrule
        Adapt Anything (Ours) & $\times$ & \multicolumn{2}{c|}{\textbf{82.5}} & \multicolumn{2}{c|}{\textbf{96.5}} & \multicolumn{2}{c|}{\textbf{98.7}} & \textbf{92.5}\\
	\hline
	\end{tabular}
	}
	\caption{Performance comparison with single-source UDA using real source data on the ImageCLEF-DA dataset. Note that we do not present the multi-source UDA results since no existing works conduct multi-source experiments on this dataset.}
	\label{table::results-ImageCLEF}
\end{table*}

\subsection{Synthetic Data Generation}
\label{text-prompt}
The diversity of the synthetic data depends on the diversity of text prompts. A simple text prompt will lead to mode collapse of synthetic data, which will further lead to limited knowledge transfer to the image classifier adapted to the unlabeled target data. To avoid this situation, it is necessary to diversify the text prompts given domain and category descriptions. To study this problem, we design three mechanisms to generate text prompts: 1) Simple-Prompt; 2) Domain-Prompt; 3) GPT-Prompt. Simple-Prompt simply uses {\tt a photo of a [Category Name]} as text prompts for text-to-image generation. Domain-Prompt adds the domain name based on Simple-Prompt, e.g., {\tt a [Domain Name] photo of a [Category Name]}. Compared with Simple-Prompt and Domain-Prompt, GPT-Prompt asks GPT to diversify text prompts given domain and category name as well as their detailed description. We can refer to Figure \ref{fig:data} for an example illustration, which asks GPT to diversify text prompts with category name {\tt Chair} and domain name {\tt Art Painting}. GPT can provide abundant text prompts for us, such as {\tt Abstract expressionist interpretation of a rocking chair} and {\tt Sketch of a vintage wingback chair}, which help synthesize diverse surrogate images for knowledge transfer. Note that the labels of the corresponding synthetic images are derived directly from {\tt [Category Name]} in text prompts without human annotation. Experiments in the following section validate the superiority of GPT-Prompt.

\renewcommand\arraystretch{1} 
\begin{table*}[t]
	\centering
        \setlength\tabcolsep{3pt}
	\resizebox{1.\textwidth}{!}{
	\begin{tabular}{l|c|ccc|ccc|ccc|ccc|c}
	\hline
        Methods & Real & Ar$\rightarrow$Cl & Pr$\rightarrow$Cl& Re$\rightarrow$Cl & Ar$\rightarrow$Pr & Cl$\rightarrow$Pr& Re$\rightarrow$Pr & Ar$\rightarrow$Re & Pr$\rightarrow$Re& Cl$\rightarrow$Re & Cl$\rightarrow$Ar & Pr$\rightarrow$Ar& Re$\rightarrow$Ar & AVG \\
        \midrule
        SymNets \cite{Zhang2019DomainSymmetricNF} & \multirow{8}{*}{\checkmark} & 47.7 & 48.8 & 52.6 & 72.9 & 71.3 & 82.7 & 78.5 & 79.5 & 74.2 & 64.2 & 64.2 & 74.5 & 67.6\\
        CDAN+E \cite{Long2017ConditionalAD} & & 54.6 & 52.3 & 57.3 & 74.1 & 72.2 & 82.8 & 78.1 & 79.1 & 74.1 & 63.0 & 61.6 & 72.3 & 68.5\\
        DCAN \cite{Li2020DomainCA}& & 54.5 & 52.7 & 59.1 & 75.7 & 74.0 & 83.5 & 81.2 & 80.6 & 76.3 & 67.4 & 67.4 & 74.1 & 70.5\\
        BNM \cite{Cui2020TowardsDA} & & 56.7 & 55.1 & 57.0 & 77.5 & 76.3 & 84.3 & 81.0 & 82.0 & 77.1 & 67.3 & 65.3 & 73.6 & 71.1\\
        FixBi \cite{Na2020FixBiBD} & & 58.1 & 57.9 & 62.9 & 77.3 & 79.5 & 86.7 & 80.4 & 81.7 & 78.1 & 67.7 & 65.8 & 76.4 & 72.7\\
        CoVi \cite{Na2021ContrastiveVS}& & 58.5 & 60.2 & 63.6 & 78.1 & 80.0 & 86.5 & 80.0 & 82.1 & 77.0 & 68.1 & 66.4 & 76.6 & 73.1\\
        CDTrans-S \cite{Xu2021CDTransCT} & & 60.6 & 56.7 & 59.1 & 79.5 & 81.0 & 85.5 & 82.4 & 84.4 & 82.3 & 75.6 & 72.5 & 77.0 & 74.7\\
        CDTrans-B \cite{Xu2021CDTransCT} & & 68.8 & 63.3 & 66.0 & 86.8 & 87.1 & 90.6 & 86.9 & 88.2 & 87.3 & 81.5 & 79.6 & 82.0 & 80.5\\
	\hline
	\hline
        Methods & Real & \multicolumn{3}{c|}{Ar+Pr+Re$\rightarrow$Cl} & \multicolumn{3}{c|}{Ar+Cl+Re$\rightarrow$Pr} & \multicolumn{3}{c|}{Ar+Pr+Cl$\rightarrow$Re} & \multicolumn{3}{c|}{Cl+Pr+Re$\rightarrow$Ar} & AVG \\
        \midrule
        SImpAl$_{50}$ \cite{Venkat2021YourCC}& \multirow{6}{*}{\checkmark} & \multicolumn{3}{c|}{56.3} & \multicolumn{3}{c|}{80.2} & \multicolumn{3}{c|}{81.5} & \multicolumn{3}{c|}{70.8} & 72.2\\
        MFSAN \cite{Zhu2019AligningDD}& & \multicolumn{3}{c|}{62.0} & \multicolumn{3}{c|}{80.3} & \multicolumn{3}{c|}{81.8} & \multicolumn{3}{c|}{72.1} & 74.1\\
        MDAN \cite{Zhu2019AligningDD}& & \multicolumn{3}{c|}{67.0} & \multicolumn{3}{c|}{81.0} & \multicolumn{3}{c|}{82.8} & \multicolumn{3}{c|}{68.1} & 74.8\\
        MDMN \cite{Li2018ExtractingRB}& & \multicolumn{3}{c|}{67.8} & \multicolumn{3}{c|}{81.4} & \multicolumn{3}{c|}{83.3} & \multicolumn{3}{c|}{68.7} & 75.3\\
        DARN \cite{Wen2019DomainAN}& & \multicolumn{3}{c|}{68.4} & \multicolumn{3}{c|}{82.8} & \multicolumn{3}{c|}{83.9} & \multicolumn{3}{c|}{70.0} & 76.3\\
        CMSDA \cite{Scalbert2021MultiSourceDA}& & \multicolumn{3}{c|}{67.7} & \multicolumn{3}{c|}{84.2} & \multicolumn{3}{c|}{83.0} & \multicolumn{3}{c|}{71.5} & 76.6\\
 
	\hline
	\hline
        Methods & Real & \multicolumn{3}{c|}{Syn$\rightarrow$Cl} & \multicolumn{3}{c|}{Syn$\rightarrow$Pr} & \multicolumn{3}{c|}{Syn$\rightarrow$Re} & \multicolumn{3}{c|}{Syn$\rightarrow$Ar} & AVG \\
        \midrule
        Adapt Anything (Ours)& $\times$ & \multicolumn{3}{c|}{\textbf{71.6}} & \multicolumn{3}{c|}{\textbf{88.4}} & \multicolumn{3}{c|}{\textbf{89.0}} & \multicolumn{3}{c|}{\textbf{84.2}} & \textbf{83.3}\\
	\hline
	\end{tabular}
	}
	\caption{Performance comparison with single- and multi-source UDA using real source data on the Office-Home dataset.}
	\label{table::results-Office-Home}
\end{table*}

\renewcommand\arraystretch{1} 
\begin{table*}[t]
	\centering
        \setlength\tabcolsep{6pt}
	\resizebox{1.\textwidth}{!}{
	\begin{tabular}{l|c|cccccccccccc|c}
	\hline
        Methods & Syn & Plane &Bcycl &Bus &Car &Horse &Knife &Mcycl &Person &Plant &Sktbrd &Train &Truck &AVG \\
	\midrule
        SWD \cite{Lee2019SlicedWD} & \multirow{6}{*}{3D} & 90.8& 82.5 &81.7& 70.5 &91.7 &69.5 &86.3 &77.5 &87.4 &63.6 &85.6 &29.2& 76.4\\
        DTA \cite{Lee2019DropTA}& & 93.7& 82.2& 85.6 &83.8& 93.0 &81.0& 90.7& 82.1& 95.1& 78.1& 86.4 &32.1 &81.5\\
        CAN \cite{Kang2019ContrastiveAN}& & 97.0 &87.2& 82.5& 74.3 &97.8 &96.2 &90.8& 80.7& 96.6 &96.3 &87.5 &59.9 &87.2\\
        FixBi \cite{Na2020FixBiBD} & & 96.1 &87.8 &\textbf{90.5} &90.3 &96.8 &95.3 &92.8 &\textbf{88.7} &97.2 &94.2 &90.9 &25.7 &87.2\\
        CoVi \cite{Na2021ContrastiveVS} & & 96.8 &85.6 &88.9 &88.6 &97.8 &93.4 &91.9 &87.6 &96.0 &\textbf{93.8} &93.6 &48.1 &88.5\\
        CDTrans-B \cite{Xu2021CDTransCT} & & 97.1 &\textbf{90.5} &82.4 &77.5 &96.6 &96.1 &93.6 &88.6 &97.9 &86.9 &90.3 &\textbf{62.8} &88.4\\
	\midrule
        Adapt Anything (Ours) & T2I & \textbf{99.0} & 88.0 & 89.4 & \textbf{92.1} & \textbf{98.4} & \textbf{97.7} & \textbf{96.8} & 76.6 & \textbf{98.0} & 93.2 & \textbf{94.7} & 52.4 & \textbf{89.7}\\
	\hline
	\end{tabular}
	}
	\caption{Performance comparison with Syn-to-Real Adaptation on the VisDA-17 dataset, which uses 3D models for source image rendering. ``Syn'' represents the manners of synthesizing fake source images. ``3D'' denotes the manner to use 3D models to render images from different angles and with different lighting conditions. ``T2I'' denotes the manner to synthesize data using text-to-image generators.}
	\label{table::results-VisDA}
\end{table*}
\subsection{Coarse-to-Fine Knowledge Transfer: Inter- and Intra-Domain Knowledge Transfer}
The synthetic data generated in the first stage serves as a bridge to transfer the knowledge from the task-agnostic text-to-image generator to the task-oriented image classifier adapted to the unlabeled target domain. After obtaining the synthetic data with labels, we design knowledge transfer into a coarse-to-fine process, where the coarse phase is designed as an Inter-Domain Knowledge Transfer and the fine phase is designed as an Intra-Domain Knowledge Transfer. As shown in Figure \ref{fig:data}, a fraction of synthetic data is inevitably mismatched with the target category, which will lead to risks of negative transfer, motivating us to propose a simple yet effective coarse-to-fine knowledge transfer method. 

\paragraph{Inter-Domain Knowledge Transfer} We can utilize the popular UDA methods to achieve Inter-Domain Knowledge Transfer, such as MCD \cite{Saito2017MaximumCD} and CDTrans \cite{Xu2021CDTransCT}, so as to align the labeled synthetic domain and the unlabeled real domain, which is formulated as follows:
\begin{equation}
    \theta = \arg\min_{\theta}\mathcal{L}_{UDA}(X, U | \theta)
\end{equation}where $X$ and $U$ denote the synthetic domain and the real domain. $\theta$ is the parameters of the target classifier, and $\mathcal{L}_{UDA}(\cdot)$ can be any UDA method.

\paragraph{Intra-Domain Knowledge Transfer} After Inter-Domain Knowledge Transfer, we propose to discard the imperfect synthetic data and exploit the unlabeled target data for self-training to achieve Intra-Domain Knowledge Transfer. First of all, we can obtain the pseudo labels for the target data with confidence scores after Inter-Domain Knowledge Transfer. We suppose that the pseudo labels with high confidence scores are reliable enough. In this way, we can transfer the knowledge from the high confident sub-domain to the unconfident sub-domain. To this end, we determine the average confidence score for each category, and use the average confidence score as a threshold to split each category into a confident part $U_{conf}$ and a unconfident part $U_{unconf}$:
\begin{equation}
    \{U_{conf}, U_{unconf}\}\xleftarrow{split} U
\end{equation}
As a result, we can treat the confident part as labeled ones and the unconfident part as unlabeled ones, and then utilize the popular semi-supervised learning (SemiSL) methods $\mathcal{L}_{SemiSL}$ to achieve Intra-Domain Knowledge Transfer, such as MixMatch \cite{Berthelot2019MixMatchAH} and FixMatch \cite{Sohn2020FixMatchSS}. In this way, $\theta$ can be further optimized via:
\begin{equation}
    \theta = \arg\min_{\theta}\mathcal{L}_{SemiSL}(U_{conf}, U_{unconf} | \theta)
\end{equation}
Overall, the entire pipeline is illustrated in Figure \ref{fig:pipeline}. Note that we do not pursue a novel method in this paper, but aim to study the feasibility of Adapt Anything. The proposed coarse-to-fine knowledge transfer is simple yet effective enough to answer the question raised in this paper.

\section{Experiments And Findings}

\subsection{Datasets}

We conduct experiments on four very popular UDA datasets, including Office-31~\cite{Saenko2010AdaptingOffice31}, Office-Home~\cite{Venkateswara2017DeepOfficeHome}, ImageCLEF-DA~\cite{Caputo2014ImageCLEF} and VisDA-17~\cite{Peng2017VisDATV}:
\begin{itemize}
    \item \textbf{Office-31} dataset contains 31 categories of office items from three different domains: Amazon(A), DSLR(D) and Webcam(W). 
    \item \textbf{Office-Home} dataset is also an office items dataset. But it contains 65 categories and four widely-used different domains: Art(Ar), Clipart(Cl), Product(Pr) and Real World(Re). 
    \item \textbf{ImageCLEF-DA} dataset uses images from three different public datasets to form three domains, namely Caltech-256(C)~\cite{Griffin2007Caltech}, ImageNet ILSVRC 2012(I)~\cite{Deng2009Imagenet} and Pascal VOC 2012(P)~\cite{Everingham2010Pascal}. For each domain, there are 12 categories and 50 images in each category.
    \item \textbf{VisDA-17} is a synthetic-to-real dataset with 12 categories. The synthetic domain contains images rendered by 3D model under different circumstances, while the real domain images are collected from MS COCO~\cite{Lin2014MicrosoftCOCO}.
\end{itemize}

\subsection{Implementation Details}

In the first stage for text-to-image generation, we first use ChatGPT (GPT-4) to generate diverse text prompts to describe a given category. The interaction template with ChatGPT is shown in Figure \ref{fig:pipeline}. We then use Stable Diffusion v2.1 to generate diverse images. For Office31, Office-Home and ImageCLEF-DA datasets, we generate 200 images for each category. And for VisDA-17 dataset, considering there are several fine-grained vehicle categories, we generate 2000 images for each category. In the second stage for inter-domain knowledge transfer, we choose four popular UDA methods, including DANN \cite{Ganin2014UnsupervisedDA}, MCD \cite{Saito2017MaximumCD}, SymNets \cite{Zhang2019DomainSymmetricNF}, and CDTrans \cite{Xu2021CDTransCT}. Specifically, we use an open-source library for Transfer Learning, named tllib~\cite{jiang2022transferability, tllib}, which has reproduced DANN and MCD. We further implemented SymNets in this library. During training on synthetic data, for each domain, we use SGD optimizer to train for 30 epochs. For CDTrans, we directly use the official code with the default hyper-parameters. To compare the influence of the network backbone, we implement CDTrans-S and CDTrans-B, respectively, where CDTrans-S is built by DeiT-Small and CDTrans-B is built by DeiT-Base \cite{Touvron2020TrainingDI}. In the third stage for intra-domain knowledge transfer, we reproduce MixMatch \cite{Berthelot2019MixMatchAH} and FixMatch \cite{Sohn2020FixMatchSS} based on the code of the second stage. To cooperate with DANN, MCD and SymNets, we use the same hyper-parameters in the third stage as those in the second stage. To cooperate with CDTrans, we adjust the learning rate to 1e-4, and the rest of the hyper-parameters in the third stage are the same as those in the second stage.

\begin{figure}[t]
    \centering
    \vskip -0.15in
    \includegraphics[width=1.\linewidth]{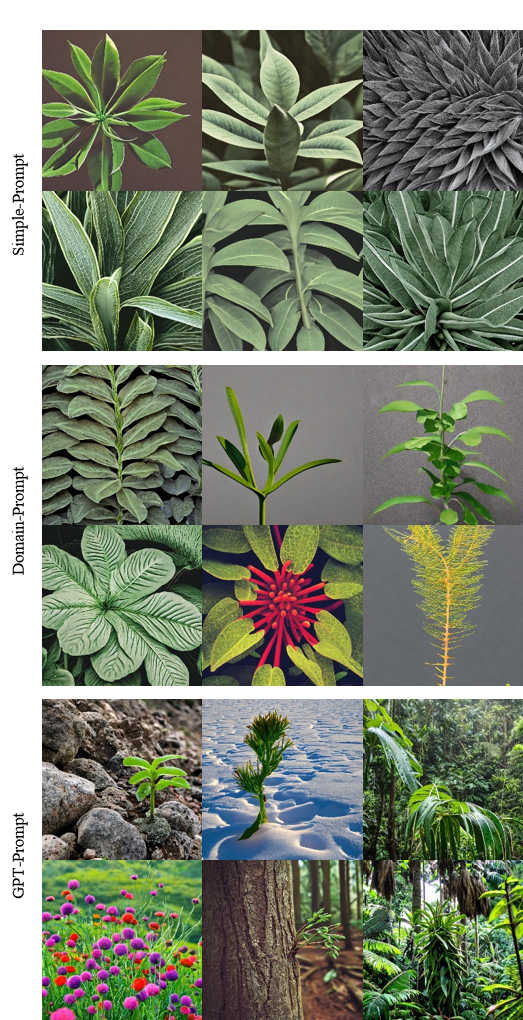}
    \caption{The visualization of synthetic source data driven by different text prompts. Category ``plant'' is used as an example.}
    \label{fig:text-prompts}
\end{figure}

\renewcommand\arraystretch{1} 
\begin{table*}[t]
	\centering
        \setlength\tabcolsep{2pt}
	\resizebox{1.\textwidth}{!}{
	\begin{tabular}{c|ccccc|ccccc|cc|cccc|c}
	\hline
        \multirow{2}{*}{Num}&\multicolumn{5}{c|}{ Stage 1} & \multicolumn{5}{c|}{ Stage 2} & \multicolumn{2}{c|}{ Stage 3} & \multicolumn{5}{c}{ Datasets}\\
	\cline{2-18}
        & SP & DP & GP & GL & SD & DANN\cite{Ganin2014UnsupervisedDA} & MCD\cite{Saito2017MaximumCD} & SN\cite{Zhang2019DomainSymmetricNF} & CD-S\cite{Xu2021CDTransCT} & CD-B\cite{Xu2021CDTransCT} & MM\cite{Berthelot2019MixMatchAH} & FM\cite{Sohn2020FixMatchSS} & Office-31 & Office-Home & ImageCLEF-DA & VisDA-17 & AVG\\
	\midrule
        {\tt 1}&\checkmark & & & & \checkmark & \checkmark & & & & & & &82.3 &70.4 &90.2 & 84.7 &81.9\\
        {\tt 2}& & \checkmark & & & \checkmark & \checkmark & & & & & & &76.4 &71.2 &85.5 &84.4 &79.3\\
        {\tt 3}& & & \checkmark& & \checkmark & \checkmark & & & & & & &75.2 &65.9 &84.7 &83.1 &77.2\\
        {\tt 4}& & & \checkmark& \checkmark & & \checkmark & & & & & & &60.1 &50.8 &86.0 &74.8 & 67.9\\
        {\tt 5}& \checkmark & & & & \checkmark & & \checkmark & & & & & &74.8 &65.4 &84.1 & 84.4 &77.1\\
        {\tt 6}& & \checkmark & & & \checkmark & & \checkmark & & & & & &79.4 &71.6 &86.3 &83.7 &80.2\\
        {\tt 7}& & & \checkmark & & \checkmark & & \checkmark & & & & & &82.4 &74.8 &88.5 &83.6&82.3\\
        {\tt 8}& \checkmark & & & & \checkmark & & & \checkmark & & & & &81.3 &65.7 &89.2 &75.8 &78.0\\
        {\tt 9}& & \checkmark & & & \checkmark & & & \checkmark & & & & &80.2 &71.6 &89.8 &73.7 &78.8\\
        {\tt 10}& & & \checkmark & & \checkmark & & & \checkmark & & & & &85.4 &74.6 &90.0 &78.6 &82.1\\
        {\tt 11}& \checkmark & & & & \checkmark & & & & \checkmark & & & &82.2 &69.9 &89.2 &86.7 &82.0\\
        {\tt 12}& & \checkmark & & & \checkmark & & & & \checkmark & & & &84.3 &74.4 &92.0 &84.5 &83.8\\
        {\tt 13}& & & \checkmark & & \checkmark & & & & \checkmark & & & &86.2 &77.7 &90.8 &87.2&85.4\\
        {\tt 14}& & & \checkmark & & \checkmark & & & & & \checkmark & & &90.2 &82.6 &92.4 &88.9&88.5\\
        {\tt 15}& & & \checkmark & & \checkmark & \checkmark & & & & & \checkmark & &81.9 &71.6 &88.8 &89.5 &82.9\\
        {\tt 16}& & & \checkmark & & \checkmark & & \checkmark & & & & \checkmark & &83.8 &76.7 &88.7 &88.2 &84.3\\
        {\tt 17}&  & & \checkmark & & \checkmark & & & \checkmark & & & \checkmark & &86.1 &75.1 &90.4 &81.9 &83.3\\
        {\tt 18}& & & \checkmark & & \checkmark & & & & \checkmark & & \checkmark & &88.2 &78.5 &91.4 &89.2 &86.8\\
        {\tt 19}& & & \checkmark & & \checkmark & & & & & \checkmark & \checkmark & &90.5 &83.3 &92.5 &89.7 &89.0\\
        {\tt 20}&  & & \checkmark & & \checkmark & & & & & \checkmark & & \checkmark & 90.5 &82.8 &92.6 &90.2 &89.0\\
	\hline
	\end{tabular}
	}
	\caption{Ablation study. Here ``SP'', ``DP'' and ``GP'' denote the Simple-Prompt, Domain-Prompt and GPT-Prompt for simplicity. ``GL'' and ``SD'' denote GLIDE \cite{Nichol2021GLIDETP} and Stable Diffusion \cite{rombach2021highresolution} models as two representatives to study the effect of different text-to-image generators. Also, ``SN'', ``CD'', ``MM'' and ``FM'' are short for SymNets \cite{Zhang2019DomainSymmetricNF}, CDTrans \cite{Xu2021CDTransCT}, MixMatch \cite{Berthelot2019MixMatchAH} and FixMatch \cite{Sohn2020FixMatchSS}, respectively.}
	\label{table::ablation}
\end{table*}
\renewcommand\arraystretch{1} 
\begin{table}[t]
	\centering
        \setlength\tabcolsep{16pt}
	\resizebox{0.48\textwidth}{!}{
	\begin{tabular}{c|c|cc}
	\hline
        UDA methods & Num & Office-31 & Office-Home\\
        \midrule
        \multirow{3}{*}{MCD \cite{Saito2017MaximumCD}} & 50 & 81.1&72.1  \\
        & 100 &81.7 &73.9  \\
        & 200 &82.4 &74.8  \\
        & 400 &82.6 &74.7  \\
        \midrule
        \multirow{3}{*}{SymNet \cite{Zhang2019DomainSymmetricNF}} & 50 &83.3 &72.8  \\
        & 100 &84.8 &73.8  \\
        & 200 &85.4 &74.6  \\
        & 400 &85.5 &74.8  \\
        \midrule
        \multirow{3}{*}{CDTrans-S \cite{Xu2021CDTransCT}} & 50 &85.4 &75.5 \\
        & 100 &85.4 &76.5 \\
        & 200 &86.2 &77.7 \\
        & 400 &86.4 &77.7 \\
	\hline
	\end{tabular}
	}
	\caption{Ablation study on the number of synthetic data.}
	\label{table::image-number}
\end{table}

\subsection{Main Results}
Without specific statement, the framework of Adapt Anything is implemented by GPT-Prompt and Stable Diffusion v2.1 in the first stage, CDTrans-B in the second stage and MixMatch in the third stage by default (we will discuss why to select this setting in the Section of ablation study). To validate whether the synthetic data from text-to-image generator can take place of real source data for domain adaptation, we compare our method with single-source domain adaptation, multi-source domain adaptation without using domain labels, as well as the syn-to-real adaptation works. We present the main results compared with the state-of-the-art UDA works using real source data in Table \ref{table::results-office31}-\ref{table::results-Office-Home} and those UDA works using synthetic data rendered from 3D models in Table \ref{table::results-VisDA}. As shown in these tables, we prove that we can achieve comparable results using the synthetic data generated from the text-to-image generator as the surrogate of the real source data. Even in most cases, our method can surpass those works using real source data by a large margin. The main reason lies in that we can generate diverse source data so as to transfer sufficient image classification knowledge to the target domains. 

However, an advantage sometimes may change into a disadvantage. In Table \ref{table::results-office31}, our results in target domains D and W fall behind most UDA works using the adaptation settings W$\rightarrow$D and D$\rightarrow$W. The main reason lies in that the domain discrepancy between D and W is much smaller than that between the synthetic data and D/W. In this way, the performances of transferring the knowledge from the synthetic data to D and W are inferior to those of transferring the knowledge from D to W or W to D. The same phenomenon also happens in traditional UDA works. As we can see in the results of the traditional single-source UDA in Table \ref{table::results-office31} (2nd-9th rows), the performances of A$\rightarrow$D (D$\rightarrow$A) and A$\rightarrow$W (W$\rightarrow$A) are worse than those of D$\rightarrow$W (W$\rightarrow$D). The reason is that the domain discrepancy between D and W is much smaller than that between A and D/W. 

Fortunately, it rarely happens in real world that the domain discrepancy between source and target domains is small enough, and this is why we surpass the UDA works using the real source data in most cases, such as the performance of transferring knowledge to domain A in Table \ref{table::results-office31} and the results in Table \ref{table::results-ImageCLEF}, Table \ref{table::results-Office-Home} and Table \ref{table::results-VisDA}. We generate the synthetic data with abundant styles so as to ensure sufficient knowledge transfer.

\subsection{Ablation Study}
In this section, we conduct extensive ablation studies for the three stages of Adapt Anything. In order to compare clearly, we summarize the results of all ablation studies in Table \ref{table::ablation} and Table \ref{table::image-number}.
 
\subsubsection{Stage 1: Diverse synthetic data}
Text prompts, text-to-image generators, and the image number of synthetic data are three most important influence factors in the first stage.

\paragraph{Text prompts} 
To evaluate the influence of text prompts, we use Simple-Prompt, Domain-Prompt and GPT-Prompt for performance comparison. The descriptions of these three manners of text prompts are demonstrated in Section \ref{text-prompt}. As shown in Table \ref{table::ablation}, comparing the experiments {\tt 5-7}, {\tt 8-10}, and {\tt 11-13}, the results of GPT-Prompt are superior to the other two methods in most cases. As shown in Figure \ref{fig:text-prompts}, the images synthesized by GPT-Prompt are apparently diverse than Simple-Prompt and Domain-Prompt. The latter two are easily trapped into a mode collapse problem. It is the diversity of the synthetic data that matters. A special case is that GPT-Prompt is inferior to the other two in experiment {\tt 1-3} in Table \ref{table::ablation}, which uses DANN for inter-domain knowledge transfer. The reason lies in the inherent drawback of DANN, which simply use GRL (Gradient Reserve Layer) for domain alignment between two domains. GRL cannot well handle source domains with rich styles.

\paragraph{Text-to-image generators}
It is intuitive that different text-to-image generators synthesize the virtual images with different quality, which takes it for granted that the ability of text-to-image generators affects the performance of Adapt Anything. To support this intuition, we choose GLIDE \cite{Nichol2021GLIDETP} as a representative to compare with the state-of-the-art text-to-image generator Stable Diffusion. As shown in the comparison of experiments {\tt 3} and {\tt 4} in Table \ref{table::ablation}. The performance of Stable Diffusion is better than GLIDE significantly. We believe it will bring more benefits to the downstream tasks as the development of text-to-image generators in the future.

\paragraph{Image number of synthetic data}
The number of the synthetic images is also one of the most important influence factor in the first stage. As shown in Table \ref{table::image-number}, the more synthetic data, the better the adaptation performance, which meets the intuition. However, the image generation stage costs a lot, which needs tremendous running time, due to the dozens of synthesis steps of Stable Diffusion. This is an inherent drawback of Stable Diffusion, which has brought the attention to the community. Also, as the increase of the image number, the performances tend to converge. For the convenience to conduct extensive experiments, we synthesize 200 images by default in Office-31, Office-Home, and ImageCLEF-DA without further increasing the image number. As for VisDA-17 dataset, considering the multiple fine-grained vehicle categories, we synthesize 2000 images in this case.

\subsubsection{Stage 2: Inter-Domain Knowledge Transfer.}
\paragraph{The effect of UDA methods}  
The selection of UDA methods is important in the second stage. We take DANN, MCD, SymNets and CDTrans as four representatives for performance comparison. Since CDTrans is the best one among them in conventional UDA settings \cite{Xu2021CDTransCT}, it is intuitive that CDTrans also performs the best among them in the setting of Adapt Anything, as shown in the experiments {\tt 3}, {\tt 7}, {\tt 10}, {\tt 13} in Table \ref{table::ablation}. Moreover, the network backbone is also an important factor of UDA methods, we perform the comparison between CDTrans-S and CDTrasns-B, which are implemented by Deit-Small and Deit-Base, respectively. As shown in the experiments {\tt 13} and {\tt 14} in Table \ref{table::ablation}, there is no surprise that CDTrans-B is better than CDTrans-S, since the network capacity of Deit-Base is larger than Deit-Small. So we recommend to use the powerful UDA method with the powerful network backbone. If we have to deploy a lightweight network in the edge devices, we recommend to apply knowledge distillation after Adapt Anything or during Adapt Anything \cite{Meng2022SlimmableDA}.

\subsubsection{Stage 3: Intra-Domain Knowledge Transfer.}
\paragraph{The importance of intra-domain knowledge transfer} As we claim in Figure \ref{fig:data}, it is inevitable that a fraction of synthetic data is mismatched with the target category, leading to noisy data in the synthetic source domain, which motivates us to conduct intra-domain knowledge transfer in the third stage. Discarding the synthetic data in the third stage for model self-evolution using target data only, the model can further increase the performance by reducing the effect of negative transfer due to the noisy data in the second stage. The experiments {\tt 14} and {\tt 19} in Table \ref{table::ablation} validates this opinion. Specifically, if the performance is much more poor in the second stage, such as using DANN in the second stage, the improvement in the third stage is much more significant. See the experiments {\tt 3} and {\tt 15} in Table \ref{table::ablation} for comparison.

\paragraph{The effect of SemiSL methods}
Similar to the second stage, the selection of SemiSL methods is important in the third stage. We use MixMatch and FixMatch as two representatives. FixMatch is expected to be better than MixMatch. However, the labeled data and the unlabeled data are splitted using the confidence metric, which are fake labeled and unlabeled parts here in the context of Adapt Anything, leading to the marginal performance difference. The results of MixMatch and FixMatch in the third stage are quite similar. For simplicity, we directly use MixMatch by default in this framework.

\subsection{Qualitative Analysis}

\begin{figure}[t]
  \centering
  \includegraphics[width=0.95\linewidth]{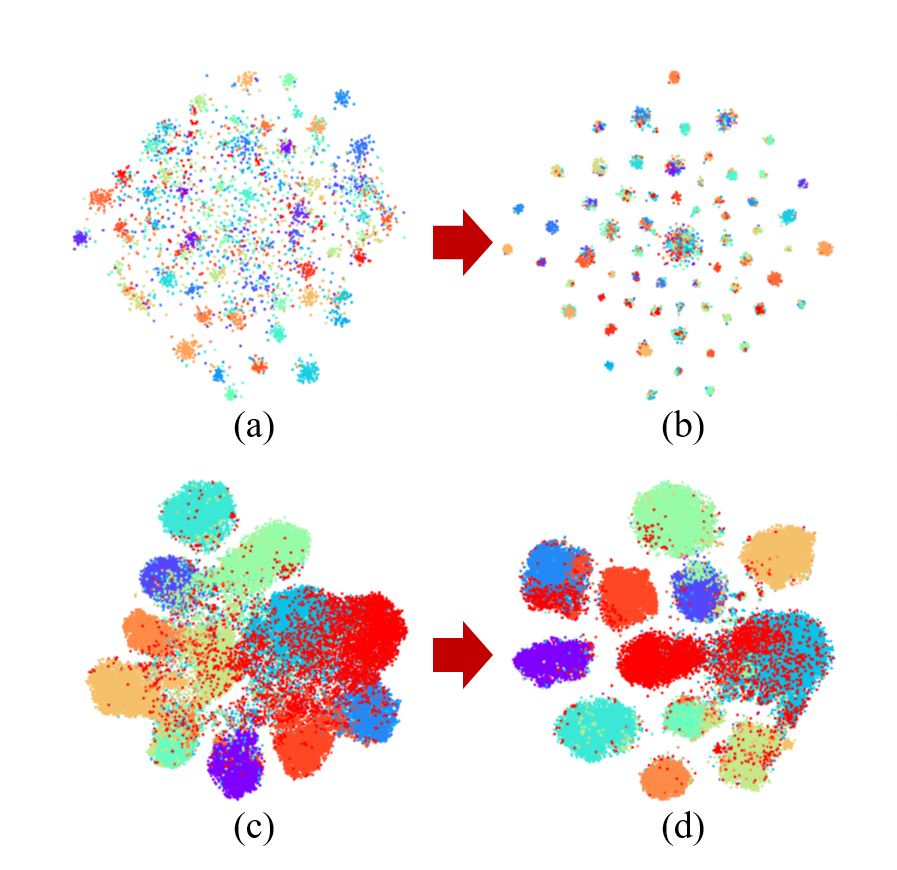}
  \caption{t-SNE feature visualization in stage 2 and stage 3, which validates the effectiveness of intra-domain knowledge transfer. (a) and (b) denotes the features in the Clipart domain of Office-Home dataset in stage 2 and stage 3, respectively. (c) and (d) denotes the counterparts in the real domain of VisDA-17 dataset. After intra-domain knowledge transfer, the features become more separable. (Each color represents each category)}
  \label{fig:tsne}
\end{figure}
\paragraph{Stage 2 vs. stage 3 via t-SNE feature visualization} To further explain why the performance in the second stage can be improved in the third stage, we visualize the features of different target images. As we can see in Figure \ref{fig:tsne}, the samples from the same category are clustered into a more compact one after intra-domain knowledge transfer. Note that the outliers of each cluster in the second stage are the unconfident samples. After the knowledge transfer from the confident ones to the unconfident ones, the unconfident samples get much closer to the confident ones, leading to a much more compact cluster in the third stage, which serves as an evidence why the performance can be further improved from the second stage to the third stage.

\section{Conclusion}
In this paper, we aim to study if a modern text-to-image diffusion model can tailor any task-adaptive image classifier across domains and categories. The answer is yes. We use the synthetic data as a bridge to transfer the knowledge from the task-agnostic text-to-image generator to the task-oriented image classifiers in the unlabeled target domains in a one-for-all manner. Using a simple yet effective Adapt Anything framework, we can achieve state-of-the-art performance on four popular domain adaptive image classification datasets, which surpass most UDA methods using real source data collected in real world or using the synthetic data rendered from 3D models. This paper demonstrates that AIGC models are mature enough to be used in real-world adaptation tasks, which we think can inspire more interesting downstream works to exploit AIGC models.

\section{Limitations And Future Works}
The idea of Adapt Anything heavily depends on whether the text-to-image generator can cover the target textual and visual concepts. It is infeasible to adapt the tasks with unknown textual and visual concepts in the text-to-image generator. This limitation in turn emphasizes the importance to develop the general text-to-image generation models.

Beyond image classification, we wonder if the remarkable text-to-image generator can tailor any task-adaptive object detection models or semantic segmentation models, which we think is a more interesting yet more challenging problem and we leave it as an open future work.

\bibliographystyle{IEEEtran}
\bibliography{main}

\end{document}